%% file: root.tex
\documentclass[letterpaper, 10 pt, conference]{ieeeconf}  

\IEEEoverridecommandlockouts                              

\usepackage{amsmath} 
\usepackage{amssymb}  
\usepackage{times}
\usepackage{graphicx}
\usepackage{multicol}
\usepackage[bookmarks=true]{hyperref}
\usepackage{xcolor}
\usepackage[ruled,vlined, linesnumbered]{algorithm2e} \SetArgSty{textup} \usepackage[noend]{algpseudocode}
\usepackage{subcaption}
\usepackage[frozencache=true,cachedir=minted-cache]{minted}
\usepackage[noadjust]{cite}
\usepackage{adjustbox}
\usepackage{appendix}
\usepackage[bottom]{footmisc}

\usepackage{svg}

\usepackage[font=small]{caption}

\title{\LARGE \bf
SPARCS: Structuring Physically Assistive Robotics for Caregiving with Stakeholders-in-the-loop
}

\author{Rishabh Madan$^{*\dagger}$, Rajat Kumar Jenamani$^{*\dagger}$, Vy Thuy Nguyen$^{\dagger}$, Ahmed Moustafa$^{\dagger}$, Xuefeng Hu$^{\dagger}$, \\Katherine Dimitropoulou$^{\ddagger}$, Tapomayukh Bhattacharjee$^{\dagger}$
\thanks{\vspace{-0.7cm}}
\thanks{*Equal contribution}
\thanks{$^{\dagger}$Rishabh Madan, Rajat Kumar Jenamani, Vy Nguyen, Ahmed Moustafa, Xuefeng Hu, Tapomayukh Bhattacharjee are with the Department of Computer Science,
        Cornell University, Ithaca, NY, USA,
        {\tt \{rm773, rj277, vtn39, aem278, xh244, tb557\}@cornell.edu}}
\thanks{$^{\ddagger}$Katherine Dimitropoulou is with Columbia University, New York City, NY, USA,
        {\tt kd2524@cumc.columbia.edu}}%
}

\begin{document}

\maketitle
\thispagestyle{empty}
\pagestyle{empty}

\begin{abstract}

Existing work in physical robot caregiving is limited in its ability to provide long-term assistance. This is majorly due to (i) lack of well-defined problems, (ii) diversity of tasks, and (iii) limited access to stakeholders from the caregiving community. We propose Structuring Physically Assistive Rob- otics for Caregiving with Stakeholders-in-the-loop (SPARCS) to address these challenges. SPARCS is a framework for physical robot caregiving comprising (i) \textit{Building Blocks}, models that define physical robot caregiving scenarios, (ii) \textit{Structured Workflows}, hierarchical workflows that enable us to answer the \textit{Whats} and \textit{Hows} of physical robot caregiving, and (iii) \textit{SPARCS-box}, a web-based platform to facilitate dialogue between all stakeholders. We collect clinical data for six care recipients with varying disabilities and demonstrate the use of SPARCS in designing well-defined caregiving scenarios and identifying their care requirements. All the data and workflows are available on \textit{SPARCS-{box}}. We demonstrate the utility of SPARCS in building a robot-assisted feeding system for one of the care recipients. We also perform experiments to show the adaptability of this system to different caregiving scenarios. Finally, we identify open challenges in physical robot caregiving by consulting care recipients and caregivers. Supplementary material can be found at \href{https://emprise.cs.cornell.edu/sparcs/}{emprise.cs.cornell.edu/sparcs}.

\end{abstract}

\input{macros/shortcuts}
\input{macros/user_operators}

\input{src/scenario}

\input{src/introduction_long}

\input{src/definitions}
\input{src/structured_workflow_shorter}

\input{src/online_tool}

\input{src/sparcs_in_practice_alternate}

\input{src/adapting_feeding}

\input{src/open_challenges}

\input{src/discussion}
\input{src/acknowledgement}




\small
\bibliographystyle{IEEEtran}
\bibliography{references}


\end{document}

%% file: macros/shortcuts.tex
\newcommand{\SPARCS}{SPARCS\xspace}
\newcommand{\BB}{\textit{Building Block}\xspace}
\newcommand{\BBs}{\textit{Building Blocks}\xspace}
\newcommand{\SW}{\textit{Structured Workflows}\xspace}

\newcommand{\FM}{\ensuremath{M_{\text{UF}}}\xspace}
\newcommand{\BM}{\ensuremath{M_{\text{UB}}}\xspace}
\newcommand{\EM}{\ensuremath{M_{\text{E}}}\xspace}
\newcommand{\RM}{\ensuremath{M_{\text{R}}}\xspace}
\newcommand{\CFM}{\ensuremath{M_{\text{CF}}}\xspace}
\newcommand{\CBM}{\ensuremath{M_{\text{CB}}}\xspace}

\newcommand{\activity}{\textit{Activity}\xspace}
\newcommand{\activities}{\textit{Activities}\xspace}
\newcommand{\compositetask}{\textit{Composite Task}\xspace}
\newcommand{\compositetasks}{\textit{Composite Tasks}\xspace}
\newcommand{\task}{\textit{Task}\xspace}
\newcommand{\tasks}{\textit{Tasks}\xspace}
\newcommand{\compositeskill}{\textit{Composite Skill}\xspace}
\newcommand{\compositeskills}{\textit{Composite Skills}\xspace}
\newcommand{\skill}{\textit{Skill}\xspace}
\newcommand{\skills}{\textit{Skills}\xspace}
\newcommand{\motorskill}{\textit{Motor Skill}\xspace}
\newcommand{\motorskills}{\textit{Motor Skills}\xspace}
\newcommand{\perceptualskill}{\textit{Perceptual Skill}\xspace}
\newcommand{\perceptualskills}{\textit{Perceptual Skills}\xspace}
\newcommand{\motor}{\textit{Motor}\xspace}
\newcommand{\perceptual}{\textit{Perceptual}\xspace}

\newcommand{\twh}{{\textit{Task Workflow for Human Caregiving}}\xspace}
\newcommand{\twr}{{\textit{Task Workflow for Robot Caregiving}}\xspace}
\newcommand{\awr}{{\textit{Action Workflow for Robot Caregiving}}\xspace}

%% file: macros/user_operators.tex

\def\rephrasek#1{\textcolor{blue}{#1}} 
\def\tnote#1{\textcolor{red}{Tapo: #1}} 
\def\rkjnote#1{\textcolor{cyan}{Rajat: #1}} 
\def\rmnote#1{\textcolor{orange}{Rishabh: #1}} 


%% file: src/scenario.tex
\vspace{-0.1cm}
\section{Caregiving is Multifaceted and Contextual}

Morgan (he/him) is a 58-year-old who sustained a brainstem stroke at 40 that made him quadriplegic and mute. He receives assistance from his partner and primary caregiver Riley (she/her), and his professional caregiver Spencer (they/them)\footnote{These are real people whom we interviewed as part of a user study. We use fictitious names/pronouns for this paper.}. Morgan's caregiving is complex and personalized. It is extremely challenging for any arbitrary person without proper training to assist him. His assistance is contextual on his functional abilities and behavior, his caregiver, and his environment: 

\noindent
\textit{1. Care Recipient Context:} When being assisted with feeding, Morgan turns to his caregiver to show his intent to have a bite. The mobility limitations in his tongue necessitate solid food to be placed around his lower left molar for chewing.

\noindent
\textit{2. Caregiver Context:} Morgan is tall. Dressing him involves lifting his limbs. While Spencer is tall and muscular and does not face difficulties with this task, Riley performs it differently because she is short and lean. 

\noindent
\textit{3. Environment Context:} Morgan lives in a house with a small bathroom. Bathing Morgan is especially challenging due to his occasional spasms. These spasms lead to involuntary movements, which can hurt him if his limbs hit the bathroom walls. This requires his caregivers to be extremely vigilant.

Many of Morgan's activities of daily living (ADLs) easily take more than an hour and pose a challenge to all other family routines. Caregiving for Morgan is very taxing for Riley and leaves her with barely any time for her own needs. Morgan is aware of the situation and wishes to do some of these activities by himself, not only to feel more independent but also to reduce Riley's burden.

\input{figures/representative}

%% file: figures/representative.tex
\begin{figure}[!t]
\centering
\includegraphics[width=\columnwidth]{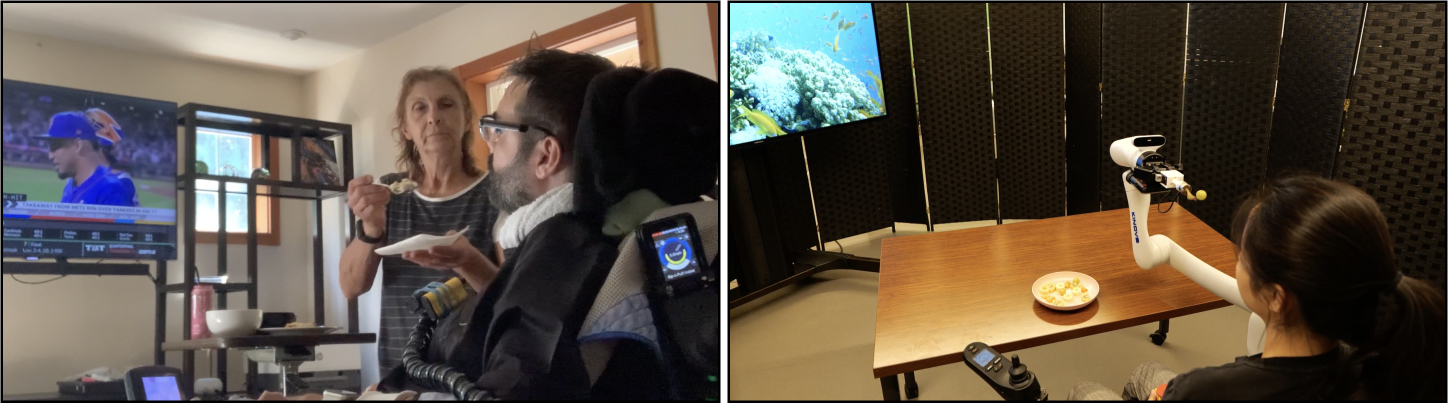}
\caption{\small 
SPARCS enables roboticists to translate human caregiving requirements to guidelines for physical robot caregiving.
}
\label{fig:representatitve_fig}
\vspace{-0.75cm}
\end{figure}

%% file: src/introduction_long.tex
\section{Introduction}
\raggedbottom
Morgan is among many others worldwide who require assistance with ADLs. Nearly $27\%$ of people living in the United States have a disability, and close to $24$ million people aged $18$ years or older need assistance with ADLs \cite{united2014americans}. Similar to Riley and Spencer, there are several family and professional caregivers who are overburdened and experience mental stress and declining quality of life \cite{gracca2013quality, dreer2007family, chio2006caregiver}. Robots have the potential to provide assistance with ADLs \cite{brose2010role} and empower people with disabilities by enhancing their independence~\cite{broekens2009assistive}, while also reducing caregiver burden. 

Existing work in physical robot caregiving is limited in its ability to consistently provide in-situ assistance to care recipients for the entire duration of an ADL. This is due to a myriad of factors such as diversity of tasks, the need for personalized assistance, and limited access to other stakeholders (care recipients, their caregivers, and occupational therapists) who have the required domain knowledge. There exists variability in problem definitions, which points to an inherent lack of common grounding in problem setup. Without a common underlying framework to systematically reason about physical robot caregiving, it is very challenging to transfer insights from one specific scenario to another.

We propose \textbf{S}tructuring \textbf{P}hysically \textbf{A}ssistive \textbf{R}obotics for \textbf{C}aregiving with \textbf{S}takeholders-in-the-loop, or \textbf{SPARCS} to address these challenges. This framework comprises \textbf{Building Blocks} (Sec. \ref{bblocks}), \textbf{Structured Workflows} (Sec. \ref{structured_workflows}), and \textbf{SPARCS-box} (Sec. \ref{sparcsbox}). Inspired by the different contexts in human caregiving, we define four \BBs,  the \textit{user}, i.e., the care recipient, their \textit{human caregiver}, their \textit{environment}, and the \textit{robot}. We build upon these blocks to form \SW, hierarchical workflows that enable us to answer the \textit{whats} and \textit{hows} of physical robot caregiving. Building these workflows requires communication between roboticists and other stakeholders who have expertise in caregiving. \textit{SPARCS-box} is a web-based platform that facilitates this dialogue.

SPARCS is a framework for building physical caregiving robots. We perform experiments that highlight the importance of \BBs, particularly user models, to build effective and personalized systems. We show SPARCS in practice by using it to (i) identify care requirements, and (ii) utilize these requirements to build a physical caregiving robot. We consider six care recipients with varying disabilities and collect data on the \BBs for ADLs that they require assistance with. We perform a user study with expert occupational therapists to build the \textit{whats} of \textit{Structured Workflows} for all the identified caregiving scenarios. This can help us translate the caregiving needs of the six care recipients into well-defined research problems for roboticists. All the data and workflows are publicly available on \textit{SPARCS-box} \cite{sparcsbox22} and can be used by roboticists to reason about the \textit{hows} for robot caregiving.  We illustrate the process of generating physical robot caregiving solutions for one of the scenarios - the activity of robot-assisted feeding for one of the six care recipients. Additionally, we demonstrate how this feeding system adapts to changes in the different building blocks. Finally, we conduct a study with care recipients and caregivers, asking them questions on general ADL assistance. We use SPARCS to examine their caregiving needs and share open challenges for the robotics community. \looseness=-1

We summarize our contributions as follows:
\begin{itemize}
    \item We propose SPARCS, a framework for physical robot caregiving. SPARCS comprises (i) \BBs, mo-\\els that define  physical robot caregiving scenarios, (ii) \SW, hierarchical workflows that enable us to answer the \textit{whats} and \textit{hows} of physical robot caregiving, and (iii) \textit{SPARCS-box}, a web-based platform that facilitates communication between all stakeholders.
    \item We highlight the importance of \BBs, particularly user models, to build effective and personalized physical caregiving robots.
    \item We collect real-world data and create \BBs for six care recipients with varying disabilities. We conduct a user study with occupational therapists and use \textit{Structured Workflows} to identify care requirements for these care recipients.
    \item We demonstrate the use of SPARCS for building a robot-assisted feeding system for one of the six care recipients. Additionally, we exhibit the adaptive nature of this system through reuse and transfer of its \SW across different \BBs.
    \item We identify open challenges in physical robot caregiving for roboticists through another user study with care recipients and caregivers.
\end{itemize}

%% file: src/definitions.tex
\section{Building Blocks of SPARCS}
\label{bblocks}
\input{figures/descriptive}

Designing a caregiving robot involves reasoning about four \textit{Building Blocks} -- the \textbf{User}, their \textbf{Human Caregiver}, their \textbf{Environment}, and the \textbf{Robot}. 
We define these blocks using functionality and behavioral user models, corresponding caregiver models, an environment model, and a robot model.

\subsection{User Functionality Model}

The \textit{User Functionality Model} \FM represents a care recipient's physical and cognitive functioning abilities. 

Physical functioning should capture body shape and structure, muscle function, joint limits, and involuntary movements. Cognitive functioning should model mental capabilities such as attention, memory, and decision-making abilities. The \textit{Human Caregiver Functionality Model} \CFM follows a similar definition, but for the caregiver.

Body shape and structure in \FM can be represented thr-\\ ough articulated human-shape models such as SMPL-X \cite{pavlakos2019expressive}. Muscle function can be incorporated using high-fidelity musculoskeletal models \cite{rcareworld, kadlevcek2016reconstructing, ryu2021functionality}. Joint limits are pose-dependent and affected by the mobility limitations of the user. They can be approximated by existing learning-based \looseness=-1
\noindent approaches \cite{gao2015user,jiang2018data} trained on user movement data. Involuntary movements have been previously simulated using parameterized offset functions \cite{erickson2020assistive}. All of the above models for physical functioning can be adapted to user measurements such as body dimensions, range of motion (ROM) \cite{reese2016joint}, and muscle strength \cite{ciesla2011manual}. Cognitive functioning of the user can be characterized using measures such as MMSE \cite{folstein1975mini} or SLUMS \cite{morley2002saint}.
The International Classification of Functioning, Disability and Health \cite{world2007international} framework defines body structures and functions, and is employed by professionals from various fields including health, rehabilitation, and community care. It can be used to model \FM more extensively. 

\subsection{User Behavioral Model}
The \textit{User Behavioral Model} \BM represents a care recipi- ent's intent and preferences, which is necessary for providing personalized care \cite{canal2021preferences}. The \textit{Human Caregiver Behavioral Model} \CBM follows a similar definition, but for the caregiver.

Intent can be identified through explicit communication or implicit inference. Explicit communication can be establish- \\ed through an interface, such as speech or a GUI \cite{tsui2008development}. Implicit inference can be made using Bayesian models \cite{javdani2018shared, jain2019probabilistic}\\ or modeling in latent space \cite{jeon2020shared}. Preferences can be global or task-environment specific \cite{canal2017taxonomy}. For example, a user may always prefer the robot to be slow-moving regardless of the scenario, whereas their preference for the level of autonomy may be task dependent. Preferences can be user-specified \cite{canal2019adapting} or modeled using data-driven methods \cite{yang2021desire}.

\subsection{Environment Model}
\label{ssec:environment_model}
The \textit{Environment Model} \EM represents physical and social information about the user's environment. Physical information constitutes the surrounding scene and the objects (including assistive devices) present in it. Social information records the environment's social context. For example, it could be an intrapersonal setting involving only the care-recipient and their
caregiver, or an interpersonal setting with family members, or a community setting \cite{bhattacharjee2019community}. 
 
Physical information is necessary for most robotic applications. Environment scenes can be represented using topological maps and semantic point clouds, whereas objects can be encoded using Unified Robot Description Format (URDF) files \cite{urdf}. Physical information can also be represented using more detailed data structures such as  3D scene graphs \cite{rosinol20203d}. Social information can be captured through theory of mind approaches \cite{scassellati2002theory} or data-driven methods \cite{ondras2022human}.
\raggedbottom
 
\subsection{Robot Model}
The \textit{Robot Model} \RM represents the hardware specifications, kinematics and dynamics, visual and collision model, and onboard sensing capabilities of the robot. It also provides access to the raw sensor data. It is important to explicitly model the robot as it affects ADL assistance \cite{bhattacharjee2019community}. 

URDFs are commonly used to represent many of the aforementioned attributes. \RM can consist of these along with other configuration files that contain information on sensors and their metadata. 

\vspace{0.2cm}
For the experiments in this paper, we initialize each \BB $M\hspace{-5pt}: \hspace{-3pt}K \hspace{-5pt}\to \hspace{-5pt} V$ as a dictionary where $K$ and $V$ are mutable sets of keys and values, respectively. $K$ contains keywords corresponding to the attributes of the \BB. $V$ contains instantiations of these attributes. For example, in case of \FM, $K$ may contain keywords like ``Active ROM Neck Flexion" or ``Passive ROM Neck Extension" with corresponding values stored in $V$\hspace{-1.5pt}.

%% file: figures/descriptive.tex
\begin{figure*}[!t]
\vspace{0.3cm}
\centering
	\begin{subfigure}[h]{2.0\columnwidth}
		\centering
		  \includegraphics[width=\linewidth]{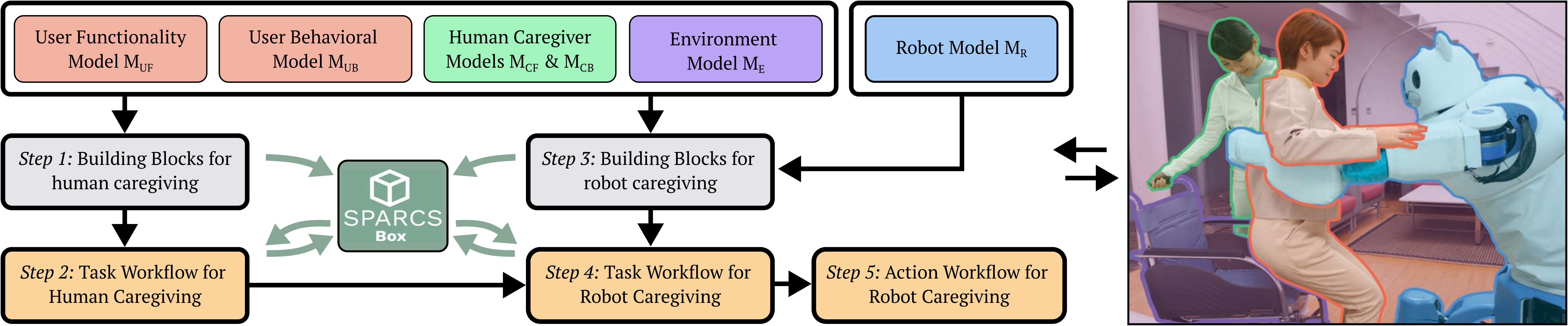}
	\end{subfigure}
\vspace{-0.1cm}
\caption{\small 
Using SPARCS involves the following steps: 1. Instantiate the \BBs for human caregiving, 2. Communicate with stakeholders using \textit{SPARCS-box} to create the \twh, 3. Instantiate the \textit{Building Blocks} for robot caregiving, 4. Propose the \twr and incorporate feedback from stakeholders using \textit{SPARCS-box}, and 5. Implement the \awr. Image shows ROBEAR from RIKEN. 
}
\vspace{-0.8cm}
\label{fig:control_policy_setup}
\end{figure*}


%% file: src/structured_workflow_shorter.tex
\section{Structured Workflows: Whats and Hows of Robot Caregiving}
\label{structured_workflows}

\vspace{-0.05cm}
Given the \BBs, how do we enable the robot to provide assistance for the entire duration of an ADL? This requires understanding the \textit{whats} that the robot must address in the caregiving scenario. We call these \textit{whats}, i.e, the set of tasks, the \twr.

While creating the \twr, roboticists must consult other stakeholders who have expertise in caregiving. However, many of these stakeholders have no experience with robotics. They cannot propose robot caregiving workflows directly. Therefore, we advocate first getting their insights on the corresponding human-human caregiving scenario. We call this the \twh. Roboticists can then use this information to propose an initial \twr, and incorporate feedback from other stakeholders on it.

Once the \twr is finalized, the focus can shift to answering \textit{how} a robot can perform the constituent tasks. We call this implementation workflow the \awr. The two \textit{Task Workflows} and this \textit{Action Workflow} are jointly referred to as \SW.

What would be a good representation for \SW? The representation must allow enough detailing to capture task-related considerations. For example, when feeding Morgan, his caregiver applies downward forces on Morgan's tongue to avoid gag reflex. At the same time, the representation should be hierarchical such that these details can be gradually uncovered as we move through various layers of abstraction. Hierarchy will also promote the reusability of similar subroutines across different caregiving scenarios. For example, the subroutine of lifting a user's leg while dressing with pants can be reused for lifting their leg during a sponge bath. The representation must also support the specification of concurrent tasks like wiping off water from the user's eyes while applying shampoo on their head. Finally, the representation must be easy to understand and update for facilitating valuable discussions among all stakeholders.

Hierarchical State Machines (HSMs) \cite{HAREL1987231} satisfy all of the above requirements. They offer an intuitive visual formalism for complex systems with various levels of abstraction. We propose to use HSMs as the representation for both \textit{Task} and \textit{Action Workflows}. Depending on the layer of abstraction, states of this machine define the routines or how they are carried out at the corresponding level of detail.
Each state has pre-conditions and post-conditions that govern transitions in and out of it. 

We define the abstractions for the two \textit{Task Workflows} in a top-down manner:

\noindent \centerline{\textit{Activity} $\rightarrow$ \textit{Composite Task} $\rightarrow$ \textit{Task}}

\noindent
An \activity is an instantiation of a given ADL. It maps to a set of \compositetasks depending on the \textit{User Functionality Model} and the initial state of the \textit{Environment Model}. \compositetasks further split into \tasks conditioned on all the \BBs. 

\newpage
In case of \twr, we map the constituent \tasks to their corresponding \textit{hows}, i.e., the \awr.
We define the abstractions for \textit{Action Workflows} as:

\noindent \hspace{1.6cm}\textit{Composite Skill} $\rightarrow$ \textit{Motor/Perceptual Skill}

\noindent 
Each \task maps to a set of \compositeskills. \compositeskill further comprises \motor and \perceptualskills. \motorskills represent the robot's ability to reason about its physical movements. \perceptualskills represent the robot's ability to interpret sensor data. Each layer of abstraction in \SW is well-scoped with respect to its parent as detailed on our website \cite{sparcsweb22}. Figure \ref{fig:representatitve_fig} shows a snippet of a \textit{Structured Workflow} for robot-assisted feeding.

\input{figures/hsm}

%% file: figures/hsm.tex
\begin{figure}[!t]
\vspace{0.15cm}
\centering
\includegraphics[width=\columnwidth]{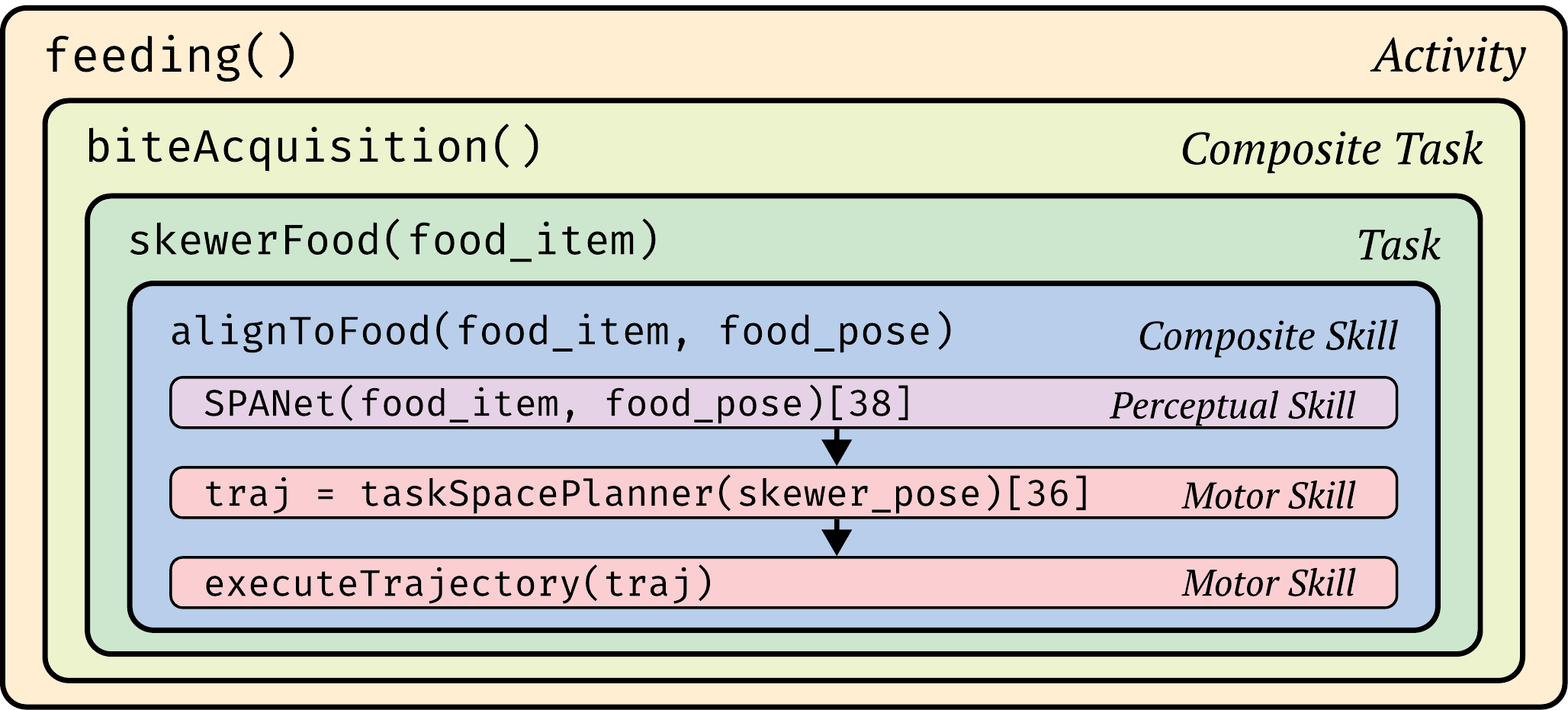}
\caption{Snippet of a \textit{Structured Workflow} for  robot-assisted feeding showing all the layers of abstraction.
}
\label{fig:representatitve_fig}
\vspace{-0.75cm}
\end{figure}

%% file: src/online_tool.tex
\vspace{-0.15cm}
\section{SPARCS-box: Bridging the Gap between Roboticists and Caregiving Community}
\label{sparcsbox}

\setlength{\skip\footins}{1pt}

Building \textit{Structured Workflows} requires roboticists to consult other stakeholders who have the required domain knowledge in caregiving. However, this is currently challenging due to limited access to stakeholders. To alleviate this problem, we propose \textit{SPARCS-box} \cite{sparcsbox22}, a web-based platform that facilitates discussion between all stakeholders. 

\textit{SPARCS-box} allows stakeholders to sign up and propose caregiving scenarios of their choice. A caregiving scenario can be specified by detailing the \activity and its \BBs. Stakeholders can collaborate on the platform to design its \twh. Roboticists can then use this information to propose the corresponding \twr. They can get feedback on it from other stakeholders on the platform. Once this workflow is finalized, roboticists can work on well-defined robot caregiving problems. They can pick \tasks from this workflow and build \skills that contribute to the \awr. We provide more details on all the functionalities of \textit{SPARCS-box} on our website \cite{sparcsweb22}.

%% file: src/sparcs_in_practice_alternate.tex
\vspace{-0.15cm}
\section{Importance of User Modeling in SPARCS}
\label{sec:user_modeling}

Physical robot caregiving requires instantiating detailed \textit{User Functionality Model} \FM and \textit{User Behavioral Model} \BM. These models help in building more effective and personalized systems. To illustrate this, we consider \textit{bite transfer} and \textit{bite sequencing} for a robot-assisted feeding activity with a wheelchair-mounted robot arm. Bite transfer involves transferring a food item from a utensil into the mouth of a care recipient. Bite sequencing involves deciding the next food item to feed a care recipient. We demonstrate how algorithms that leverage user-informed \textit{a)} \FM for bite transfer and \textit{b)} \BM for bite sequencing perform better than baselines.

\subsection{Leveraging \FM for Bite Transfer}
\label{ssec:muf_bite_transfer}
To show the effect of \FM for bite transfer, we select the user Natalia (Table \ref{tab:six_avatar}) for our experiments.
\input{figures/bite_transfer_performance_comparison}

\textit{Experiment setup:} Our bite transfer policy transfers a food item from its initial fixed position facing Natalia to near her mouth. Her head pose is denoted by $h_{user}$. We use active ROM data for her neck to instantiate the manifold $H_{\FM}$ of attainable head poses. We denote the set of all head poses attainable by an individual with full neck mobility by $H_{all}$. We consider the test set spanning all possible $h_{user}$ $\in$ $H_{\FM}$.

\textit{Methods:} 
We compare three bite transfer control policies: (i) Fixed \cite{gallenberger2019transfer} which executes a trajectory to an assumed fixed head pose $h_{fixed} \in H_{\FM}$, (ii) Baseline which uses $H_{all}$, and (iii) \FM-informed which uses $H_{\FM}$.

\FM-informed begins with sampling a set $H_{cand}$ comprising candidate head poses from \FM where the bite transfer can happen. $H_{cand}$ also includes $h_{user}$ and $h_{fixed}$. These poses are ordered according to relative angle from $h_{user}$. The algorithm sequentially iterates over each sample in $H_{cand}$. For each head pose sample $h_{cand}$ $\in H_{cand}$, the corresponding end effector goal pose $x_{goal}$ is found. 
Natalia aligns her mouth opposite to the fork while taking the food item. Thus, $x_{goal}$ is assumed to be at a fixed transform to $h_{cand}$.
We then use task-space-region planner \cite{berenson2011task} to find a collision-free trajectory from the initial configuration of the robot to $x_{goal}$. If the trajectory is successfully found, it is returned. If no trajectory is found even after iterating over all the samples, the algorithm terminates.

Baseline follows a similar approach. However, as it does not have access to $H_{\FM}$ or $h_{fixed}$, it instead constructs $H_{cand}$ using $h_{user}$ and head poses sampled from $H_{all}$.

\textit{Metrics:} Performance of the above methods is averaged over ten seeds for the test set of sampled $h_{user}$ and evaluated on two metrics. (1) \textit{Success rate}, which is the percentage of test samples where the bite transfer control policy can generate a feasible trajectory. A trajectory is said to be feasible if it is collision-free and reaches a pose that results in successful bite transfer. (2) \textit{Relative angle} by which a user has to move their neck to take the bite off the fork.

\textit{Results:} Among the compared policies, \FM-informed can always successfully find a feasible control policy and requires the smallest neck movement (Fig. \ref{fig:bite_transfer_performance_comparison}). Our experiments illustrate that access to user-informed \FM leads to more efficient physical robot caregiving.

\subsection{Leveraging \BM for Bite Sequencing}
We show the effect of modeling \BM for bite sequencing on user satisfaction. Among a set of candidate models for capturing \BM, we perform a user study to identify the model that has the highest overall user satisfaction.
\label{ssec:mub_bite_sequencing}
\input{figures/bite_sequencing_performance_comparision}

\textit{Experiment setup:} We consider a meal consisting of three bites each of four unique food items -- banana, kiwi, grape, and carrot.  
Assuming the user eats one item at a time and consumes all the food items, the bite sequence for a meal is an ordered sequence of these food items. We perform a user study with 14 participants. For each participant, we initially collect two data points. First, an affinity score $\in [1, 5]$ for each food item that implies how much they like it. Second, their high-level eating preference among (a) \textit{I would save my favorite food for the last}, (b) \textit{I would eat my favorite food first}, and (c) \textit{I prefer to mix and match}. We then ask the participants to record their bite sequences for six meals. Using this data, the goal is to learn a preference model as part of \BM and use it for predicting the bite sequence during the seventh meal.

\textit{Methods:} Our approach comprises (i) generating simulated bite sequences for a user using their affinity score and high-level eating preference, (ii) training a hidden Markov Model (HMM)
with discrete emissions using the simulated data, and (iii) updating the model using the six bite sequences obtained from the user. We use HMM-simulated (\texttt{HS}) to refer to the model obtained after (ii) and HMM-online (\texttt{HO}) for the final updated model after (iii). We compare with a baseline that randomly selects food items.

\textit{Metrics:} Performance of the above methods is evaluated on two metrics. (1) \textit{Prediction accuracy} during the seventh meal. This is calculated by presenting the predictions from all the methods as choices to the participant and comparing the predictions with the item selected by the participant. (2) For each method, we generate a bite sequence assuming the participant always chooses the item predicted by that method, and ask them to provide a \textit{satisfaction rating} $\in [1,5]$ for the generated sequence. 

\textit{Results:} We perform Kruskal-Wallis H-tests and Tukey HSD post-hoc tests and observe that \texttt{HO} significantly outperforms other methods in both accuracy and satisfaction rating (Fig. \ref{fig:bite_sequence_performance_comparison}). This user study demonstrates that modeling the user preferences for bite sequencing and adapting it to user data leads to improved satisfaction. Refer to our website \cite{sparcsweb22} for more details on the user study.

\section{SPARCS in Practice: Identifying Care Requirements and Building a Caregiving Robot}

SPARCS is a framework for building physical caregiving robots. Using SPARCS involves the following steps (Fig. \ref{fig:control_policy_setup}): \\
A. Identifying care requirements: 
\begin{enumerate}
    \item[1.] Instantiate \BBs for human caregiving.
    \item[2.] Communicate with stakeholders using \textit{SPARCS-box} to create \twh.
\end{enumerate}
B. Building a physical caregiving robot:
\begin{enumerate}
    \item[3.] Instantiate \BBs for robot caregiving.
    \item[4.] Propose \twr and inco- rporate feedback from stakeholders using \textit{SPARCS-box}.
    \item[5.] Implement \awr.
\end{enumerate}
\noindent
We demonstrate each of these steps in this section. For steps 1 and 2, we consider six care recipients with varying functional abilities (Table \ref{tab:six_avatar}). We instantiate various caregiving scenarios for each of these care recipients and identify their care requirements (Sec. \ref{ssec:identifying_care_requirements}). For steps 3, 4, and 5, we consider one of the above identified scenarios -- robot-assisted feeding of bite-sized food items to Natalia while she is watching television (Sec. \ref{ssec:building_caregiving_robot}).

\subsection{Identifying Care Requirements for Six Care Recipients}
\label{ssec:identifying_care_requirements}

We consider six care recipients with different functional abilities to capture variability in care requirements (Table \ref{tab:six_avatar}). 

\begin{table}[b]
\centering
\vspace{-0.5cm}
\caption{\small{Information for the six care-recipients. We use initials of \textbf{F}eeding, \textbf{D}ressing, \textbf{B}athing, and \textbf{T}ransferring to denote their need for assistance with these ADLs.}}
\vspace{-0.1cm}
\label{tab:six_avatar}
\begin{tabular}{ccc}
\hline
Identifier & Cause of Disability & Needs Assistance              \\ \hline
Morgan (he/him)   & Brainstem Stroke  & F, D, B, T \\ 
Jose (they/them)   & Spinal Cord Injury (C1-C3) & F, D, B, T \\
Natalia (she/her) & Spinal Cord Injury (C4-C5) & F, D, B, T \\
Daniel (he/him)   & Spinal Cord Injury (C6-C7) & D, T \\
Kim (she/her) & Cerebral Palsy & D, B, T \\
Karan (he/him)   & Left-side Hemiplegia & D, T \\ \hline
\end{tabular}
\end{table}

\input{figures/data_collection}

\vspace{0.1cm}
\noindent \textit{{\textbf{Step 1.} Instantiate Building Blocks for Human Caregiving:}}

For each of the six care recipients, we instantiate corresponding \textit{User Functionality Model} \FM, \textit{User Behavioral Model} \BM, \textit{Caregiver Models} (\CFM and \CBM), and \textit{Environment Model} \EM. We collect clinical data for information on \FM (Fig. \ref{fig:data_collection_ROM}) -- body dimensions, weight, active and passive ROM \cite{reese2016joint} and manual muscle testing \cite{ciesla2011manual} -- along with textual descriptions, and videos of non-disabled medical professionals simulating the functional abilities of these care recipients. For each care recipient, we look at ADLs -- feed-\\ ing, dressing, bathing, and transferring. Among these ADLs, we identify scenarios they require assistance with based on their functional abilities. We identify a total of 19 caregiving scenarios and provide the corresponding \BBs.

\vspace{0.1cm}
\noindent \textit{{\textbf{Step 2.} Create Task Workflow for Human Caregiving:}}

One should use \textit{SPARCS-box} to create \twh. For this paper, we conduct a user study interviewing 9 occupational therapists (2 male; 7 female), between the ages of 27 and 51. Through this study, we record \textit{Task Workflows} for the 19 caregiving scenarios.
\vspace{0.2cm}

The \BBs and above workflows for all the identified caregiving scenarios are publicly available on \textit{SPARCS-box}. This information can be leveraged by roboticists to build robots that can address the identified care requirements.

\subsection{Building a Physical Caregiving Robot}
\label{ssec:building_caregiving_robot}

In this section, we focus on the scenario of feeding Natalia while she is watching television.

\vspace{0.1cm}
\noindent \textit{{\textbf{Step 3.} Instantiate Building Blocks for Robot Caregiving:}}

We instantiate the \textit{Building Blocks} for the above scenario from the data collected in \textit{Step 1} and include our \textit{Robot Model} in it.

\noindent
- \textit{Environment Model} \EM: Natalia is sitting at the dining table on a wheelchair with a robot arm mounted beside its right armrest. The dining table is in her living room in-front of the television. There is a plate in front of her, consisting of solid bite-sized food items. This scene is captured using a pointcloud obtained from an RGBD camera. Objects in this scene are represented using URDFs \cite{urdf}. The shape and pose of Natalia's head is represented using FLAME \cite{feng2021learning}.

\noindent
- \textit{User Functionality Model} \FM: Natalia has complete paralysis in all limbs. She has partial mobility in her neck. We use the data collected in \textit{Step 1} to represent her \FM.

\noindent
- \textit{User Behavioral Model} \BM: Natalia prefers the robot to autonomously take decisions without any input from her, and shows her intent of taking a bite by opening her mouth. She tends to focus on the television while eating and thus favors making minimal neck movements to transfer the food from the fork to her mouth. Among the set of food items, she has a preferred order for eating the food items. She expects the robot to learn this ordering. We use HMM-online (Sec. \ref{ssec:mub_bite_sequencing}) to model this preference.

\noindent
- \textit{Robot Model} \RM: We use the Kinova Gen3 6 degrees-of-freedom (DoF) robot arm with a Robotiq 2F-85 gripper, holding a custom fork fitted with an ATI Force/Torque sensor.

\vspace{0.1cm}
\noindent \textit{{\textbf{Step 4.} Create Task Workflow for Robot Caregiving:}}

We build the \twr using the \twh identified in \textit{Step 2}. This \textit{Activity} can be broken down into two \textit{Composite Tasks}: (i) \textit{Bite Acquisition}: acquiring a food item from the plate, and (ii) \textit{Bite Transfer}: transferring this food item into the mouth of the care recipient. 
Natalia's \BM specifies that she prefers the caregiver to feed her without any input. Thus, \textit{Bite Acquisition} comprises the \textit{Task} of the robot autonomously moving above the plate and then skewering the required food item. According to Natalia's \FM, she has partial mobility in her neck. She can eat the food item off the fork if the robot brings it to a region within her functional abilities. \textit{Bite Transfer} begins with the robot moving to a fixed pose in front of Natalia such that her head is visible to its camera. According to her \BM, the robot waits for her to open her mouth. Once it is open, the robot brings the food item near her mouth, ensuring minimal neck movement for transfer. Natalia then leans towards the food item and takes it off the fork.

\input{figures/bite_transfer_trajectories_comparison}

\vspace{0.1cm}
\noindent \textit{{\textbf{Step 5.} Implement Action Workflow for Robot Caregiving:}}

We demonstrate the \awr with an individual simulating Natalia. For \textit{Bite Acquisition}, we deploy SPANet \cite{feng2019robot} for selecting how to skewer a food item from a set of candidate skewering actions. For \textit{Bite Transfer}, we deploy the \FM-informed policy (described in Sec. \ref{ssec:muf_bite_transfer}). The demonstration of our robot feeding an individual simulating Natalia can be seen in Fig. \ref{fig:natalia_tv_feeding} and found on our website \cite{sparcsweb22}.

%% file: figures/bite_transfer_performance_comparison.tex
\begin{figure}[!b]
\vspace{-0.5cm}
\centering
	\begin{subfigure}[h]{0.49\columnwidth}
		\centering
\includegraphics[width=\linewidth]{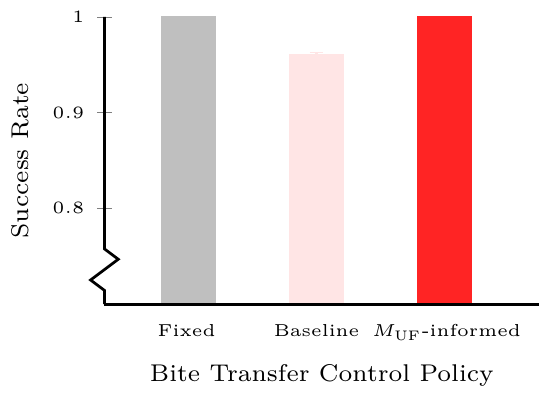}\
	\end{subfigure}
	\hfill
	\begin{subfigure}[h]{0.49\columnwidth}
		\centering
		 \includegraphics[width=\linewidth]{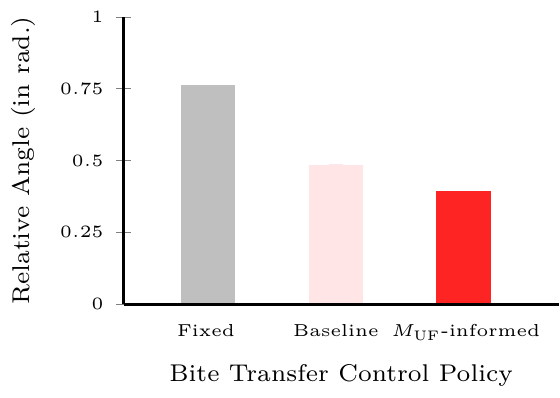}
	\end{subfigure}
\vspace{-0.2cm}
\caption{\small{Fixed and \FM-informed policies have a high bite transfer success rate, while the latter also results in minimal neck movement.}}
\vspace{-0.1cm}
\label{fig:bite_transfer_performance_comparison}
\end{figure}

%% file: figures/bite_sequencing_performance_comparision.tex
\begin{figure}[!b]
\vspace{-0.5cm}
\centering
	\begin{subfigure}[h]{0.48\columnwidth}
		\centering
		  \includegraphics[width=\linewidth]{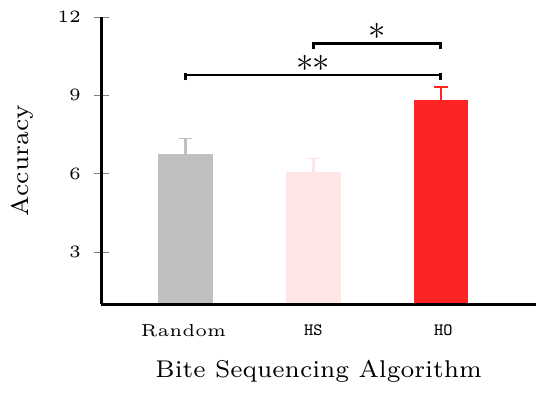}\
	\end{subfigure}
	\hfill
	\begin{subfigure}[h]{0.48\columnwidth}
		\centering
		 \includegraphics[width=\linewidth]{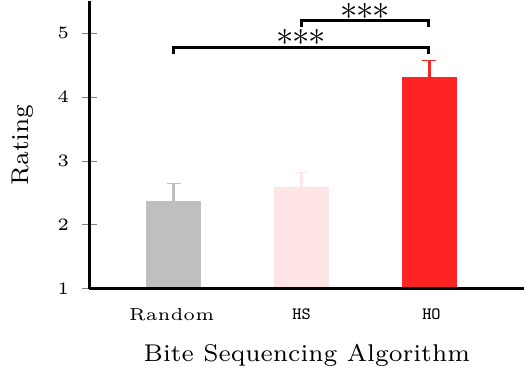}
	\end{subfigure}
\vspace{-0.2cm}
\caption{HMM-online \texttt{HO} outperforms other methods in both accuracy and satisfaction rating. $\ast$, $\ast\ast$, $\ast$$\ast\ast$ denote statistically significant differences with $p_{0.05}$, $p_{0.005}$, $p_{0.0005}$ respectively.}
\label{fig:bite_sequence_performance_comparison}
\end{figure}

%% file: figures/data_collection.tex

	
	


\begin{figure}[b]
\vspace{-0.3cm}
\centering
\includegraphics[width=\columnwidth]{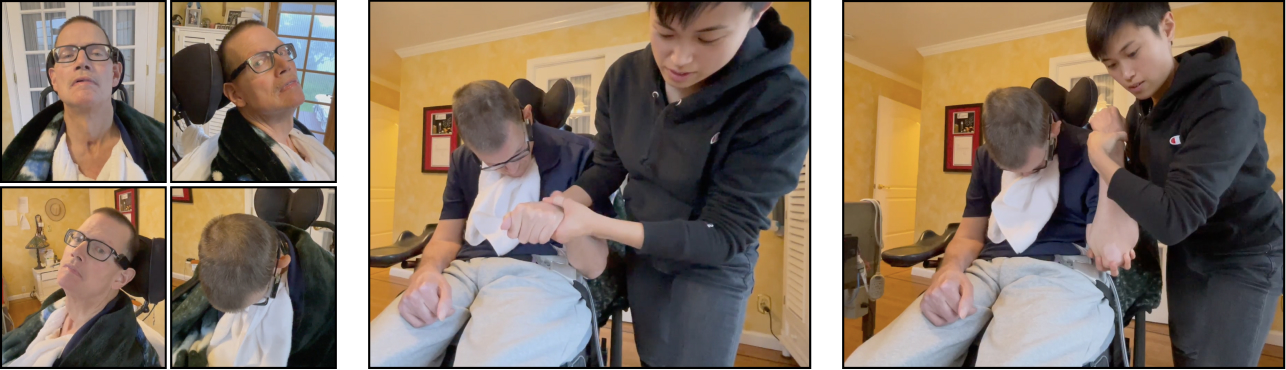}
\vspace{-0.3cm}
\caption{Data Collection for Morgan's \FM, \textit{left to right}: Active Range of Motion, Passive Range of Motion, Manual Muscle Testing}
\label{fig:data_collection_ROM}
\end{figure}

%% file: figures/bite_transfer_trajectories_comparison.tex


\begin{figure}[t]
\vspace{0.15cm}
\centering
\includegraphics[width=\columnwidth]{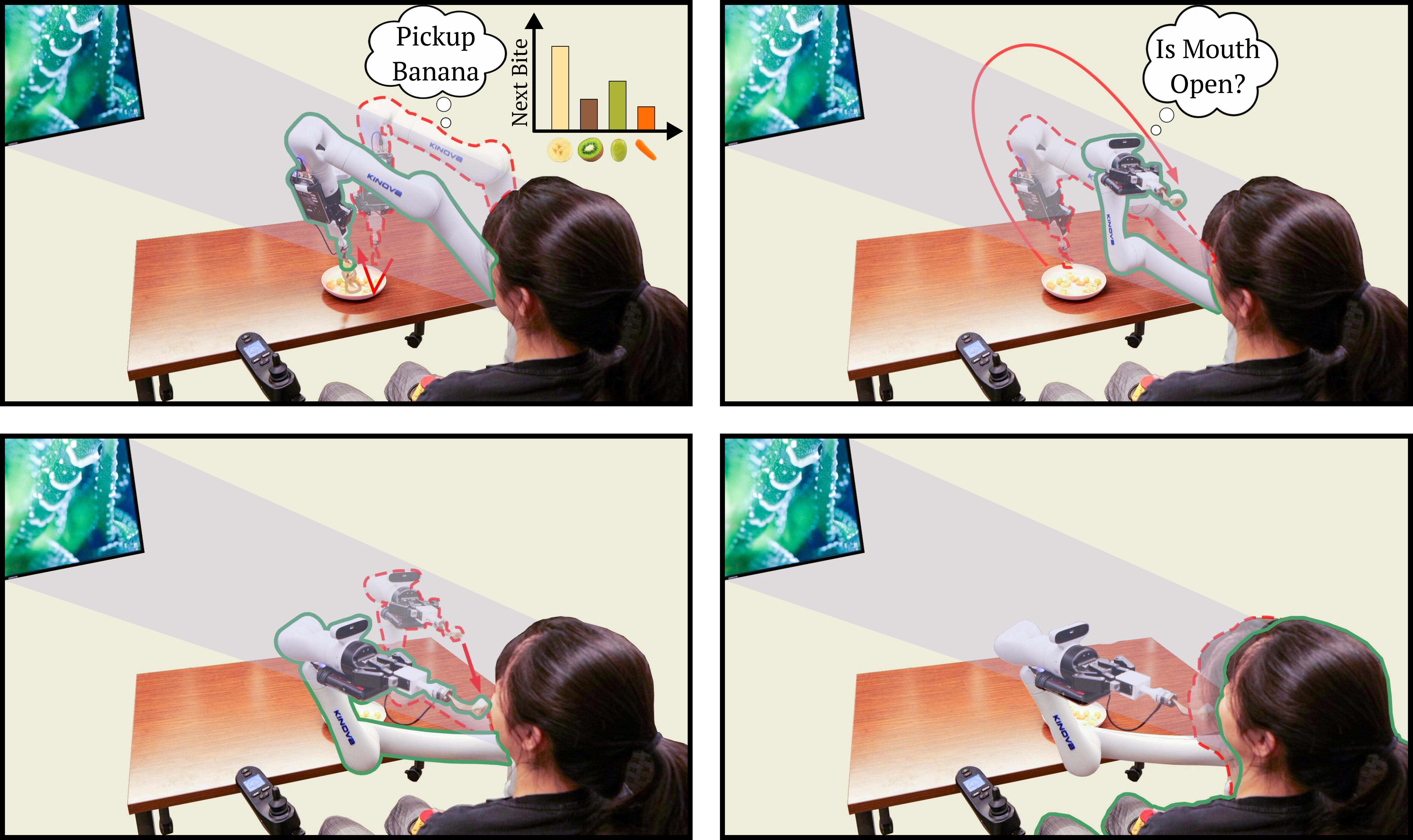}
	\caption{Robot assisted-feeding for an individual simulating Natalia while she watches TV. \textit{Top left}: Robot performs \textit{Bite Acquisition} using SPANet \cite{feng2019robot}. \textit{Top right}: It enters \textit{Bite Transfer} and waits for the user to open mouth. \textit{Bottom left}: It moves close to Natalia's mouth. \textit{Bottom right}: Natalia leans forward and takes the bite.}
	\label{fig:natalia_tv_feeding}
 \vspace{-7mm}
\end{figure}

%% file: src/adapting_feeding.tex
\section{Adapting Robot-Assisted Feeding to Different Caregiving Scenarios}
\label{ssec:building_blocks_raf}

In this section, we perform experiments to exhibit the adaptability of our system. We show how robot caregiving scenarios instantiated using SPARCS allow us to reuse and transfer insights across differing users, environments, and robot models.

\subsection{User Functionality and Behavioral Models.} 

We adapt the \SW for Natalia to Jose (Table \ref{tab:six_avatar}) while keeping all other Building Blocks fixed:

\noindent
-  New \FM: Jose has severe neck mobility limitations with partial neck rotation ROM. Unlike Natalia, they require their caregiver to place a food item inside their mouth cavity for \textit{Bite Transfer}.

\noindent
- New \BM: Unlike Natalia, Jose shows their intent for wanting a bite by turning towards the robot, and expresses their consent for the food item to be placed inside their mouth by opening it. They prefer the food item to be placed one-third inside their mouth cavity. Similar to Natalia, they expect the robot to learn their preferred bite sequence.

\noindent
In the \SW for this scenario, while \textit{Bite Acquisition} remains similar, the \textit{Task} breakdown for \textit{Bite Transfer} changes. \textit{Bite Transfer} begins with the robot moving to a fixed pose in-front of Jose such that their head is visible to its camera. In accordance with their \BM, the robot waits for them to turn towards the robot and then moves the food item to in-front of their mouth. Once Jose's mouth is open, the robot moves the food item inside their mouth cavity to successfully transfer the bite. The demonstration of our robot feeding an individual simulating Jose can be seen in Fig. \ref{fig:bt_jose} and found on our website \cite{sparcsweb22}.

\input{figures/bite_transfer_jose}

\subsection{Robot Model.}
We consider a different \textit{Robot Model} \RM while keeping the same user and environment models:

\noindent
- New \RM: We use the Kinova Gen3 7-DoF robot arm with a Robotiq 2F-85 gripper, holding a custom fork fitted with an ATI Force/Torque sensor.

\noindent
As the robot has a similar morphology to the one considered in Sec. \ref{ssec:building_caregiving_robot}, the \SW remain the same. However, the change in its dimensions and DoF affect its performance in \textit{Bite Transfer} as shown in Table II. We observe that the new \RM performs better than the previous model which could be an effect of having better manipulability.

\begin{table}[H]
\centering
\vspace{-0.3cm}
\caption{$M_R$ comparision (averaged over three seeds)}
\vspace{-0.2cm}
\label{tab:robot_model_comparision}
\begin{tabular}{ccc}
\hline
Robot Arm & Success Rate & Relative Angle (in rad.)  \\ \hline
Kinova Gen3 6-DoF   & 1.0 & 0.3996 $\mp$ 0.0018 \\
Kinova Gen3 7-DoF   & 1.0 & \textbf{0.3496 $\mp$ 0.0008}\\
\hline
\end{tabular}
\vspace{-0.2cm} 
\end{table}

\subsection{Environment Model.} 

We consider a different \textit{Environment Model} \EM while keeping the same user and robot models:

\noindent
- New \EM: Natalia is sitting at her dining table on a wheelch-\\air with a robot arm mounted beside its right armrest. There is a plate in-front of her, consisting of solid bite-sized food items. She is dining along with her friends in a social setting. 

\noindent
While dining in this social setting, Natalia prefers the robot to bring the food item to a fixed position on her right side for \textit{Bite Transfer}. This is to avoid any form of social distraction due to the robot motion. The demonstration of our robot feeding an individual simulating Natalia in the new \EM can be seen in Fig. \ref{fig:bt_social} and found on our website \cite{sparcsweb22}.ß

\input{figures/bite_transfer_natalia_social}

%% file: figures/bite_transfer_jose.tex
\begin{figure}[t]
\vspace{0.15cm}
\centering
\includegraphics[width=\columnwidth]{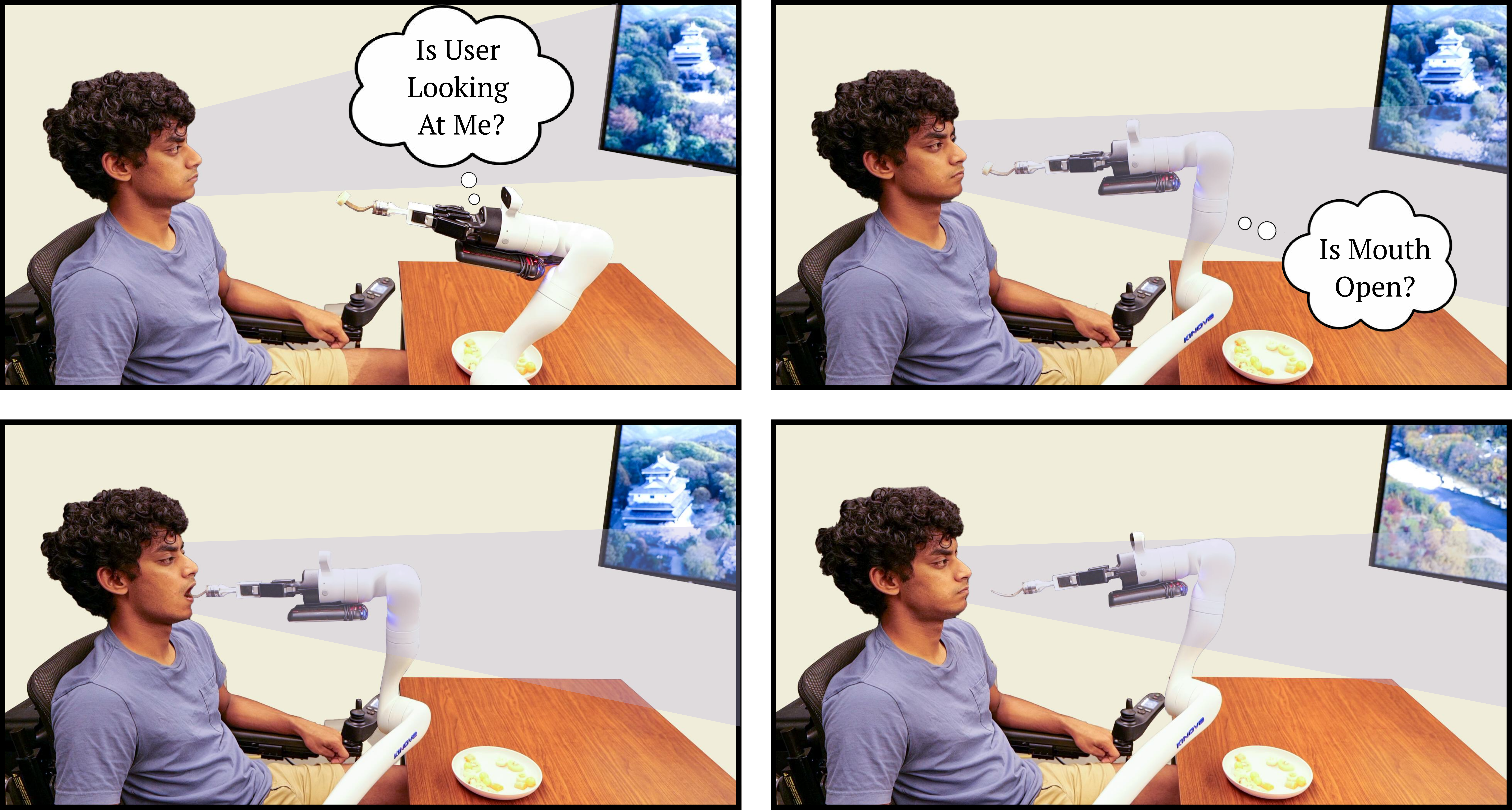}
	\caption{Robot-assisted feeding for an individual simulating Jose. \textit{Top left}: Jose watches television. \textit{Top right}: They show the intent of taking a bite by looking at the robot. \textit{Bottom left}: Robot performs an inside-the-mouth transfer due to Jose's mobility limitations. \textit{Bottom right}: Robot retracts leaving food item inside the mouth.}
	\label{fig:bt_jose}
 \vspace{-7mm}
\end{figure}


%% file: figures/bite_transfer_natalia_social.tex
\begin{figure}[t]
\vspace{0.15cm}
\centering
\includegraphics[width=\columnwidth]{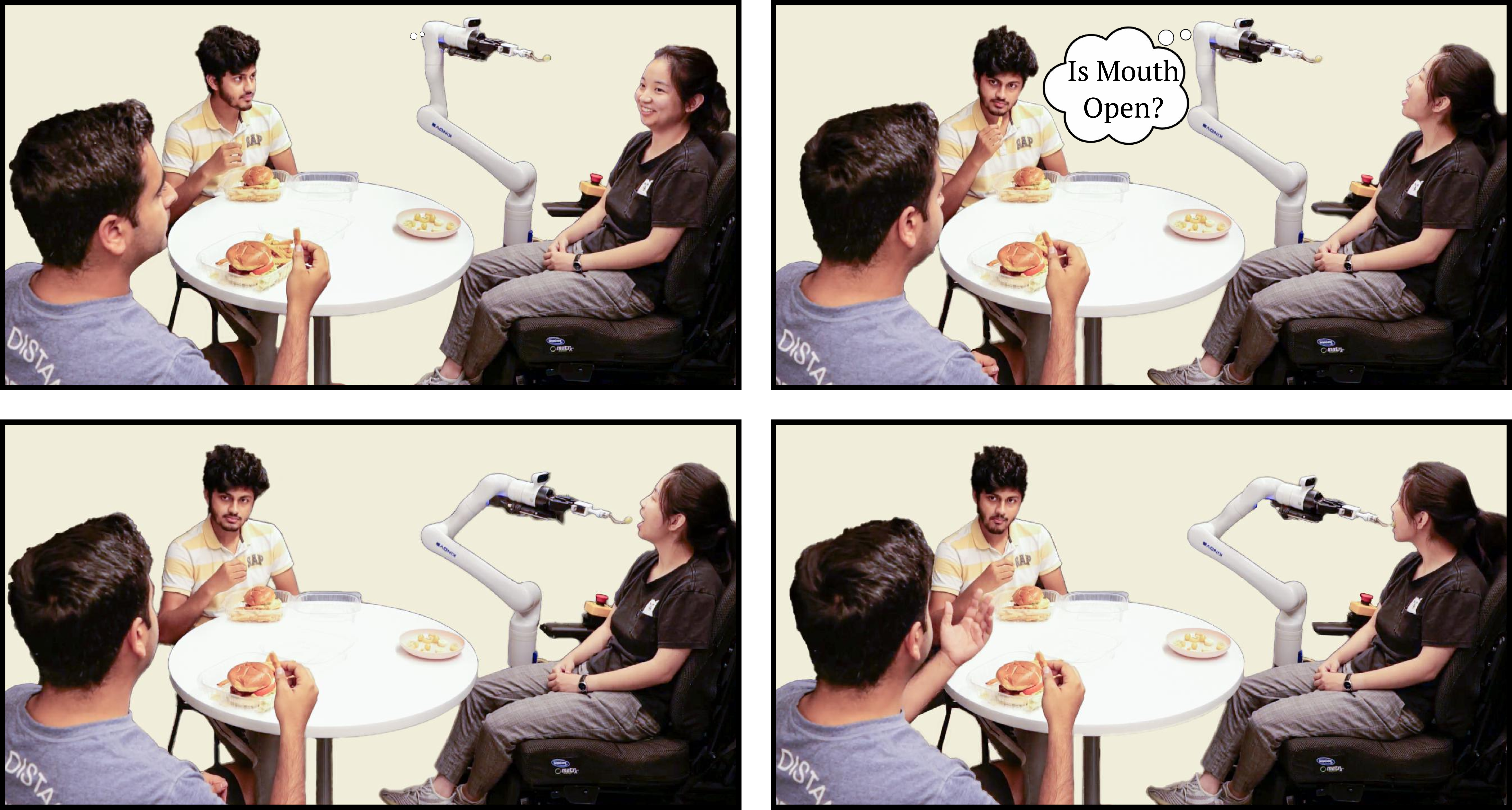}
	\caption{Robot-assisted feeding for an individual simulating Natalia in a social dining scenario. \textit{Top left}: Natalia having a conversation. \textit{Top right}: In this \EM, Natalia faces the robot and opens her mouth. \textit{Bottom left}: Robot moves close to her mouth. \textit{Bottom right}: Natalia leans forward and takes the bite.}
	\label{fig:bt_social}
 \vspace{-0.7cm}
\end{figure}


%% file: src/open_challenges.tex
\section{Open Challenges}

As more roboticists get involved in physical robot caregiving, it is important to identify open problems that one can work on. Therefore, we reached out to care recipients and \looseness=-1

\noindent
caregivers to better understand the critical factors that affect caregiving for ADLs. We used SPARCS to create caregiving scenarios around them and obtained recommendations on ADL assistance. We conducted a user study with 8 care recipients (6 male; 2 female) and 2 caregivers (both female), between the ages of 21 and 60. All data collection and user studies in this paper were approved by the Cornell University Institutional Review Board. Here we list some key open challenges:

\textbf{Feeding.} While there has been a lot of work on robot-assisted feeding, many challenges still need to be addressed for real-world applicability. Through our user study, we identify four unique challenges in feeding: (i) determining the level of autonomy because, it is not only personalized to a user but also to different tasks within feeding, (ii) acquiring a variety of food items with different physical characteristics across shape, size, texture, compliance, etc., (iii) designing the embodiment such that it can potentially perform other tasks while still being well suited for feeding, and (iv) determining when to switch between human supervision and autonomous control with our users preferring to switch to human supervision at the tail-end of a task (e.g. precisely aligning to a food item for \textit{Bite Acquisition}) for finer control.

\textbf{Dressing and Bathing.}
Shared autonomy methods purely based on joystick-based control have been considered viable for controlling caregiving robots. However, they may not be feasible for many tasks within dressing and bathing. For example, when dressing a user with a t-shirt, the user may not be able to see around and control the joystick due to occlusion. Thus, a significant challenge here is to build caregiving robots that allow multiple modes of interfaces to control the robot. Participants also highlighted that dressing the lower body is more complex than the upper body. In most cases, lower body dressing is performed while the care recipient is lying on their bed. Dressing them in this state requires high payload capacity and reasoning under partial observability. These requirements make the physical interaction in this task particularly challenging. Also, bathing and dressing tasks would conceivably require a robot to collaborate with the user and their caregiver(s), and task allocation during physical caregiving is a major challenge.

\textbf{Transferring.} 
At first glance, though transferring may seem to require a robot with high payload capacity, our users highlight that existing robot hardware can still be used. This is possible by collaborating with human caregivers to operate assistive devices such as hoyer lift, sit-to-stand lift, sliding board, etc. A major challenge is to design intelligent collaboration policies, and enable seamless integration between these robots and other assistive devices. As pointed out by a care recipient, transferring can begin with coarsely moving them to a bed or a wheelchair using an assistive device such as a hoyer lift. The robot can then be used for finer limb-repositioning to successfully complete this ADL.

%% file: src/discussion.tex
\section{Discussion}

We introduced SPARCS, a framework for physical robot caregiving. SPARCS enables roboticists to translate real-world care requirements into guidelines for physical robot caregiving. Occupational therapists use similar frameworks when designing caregiving interventions. Compared to these frameworks, we define the components of SPARCS by grounding them in robotics. We release \textit{SPARCS-box}, which allows roboticists to communicate with stakeholders from the caregiving community. In the future, we intend to improve \textit{SPARCS-box} to make it more accessible and add features that incentivize stakeholders to use this platform. 

SPARCS currently defines \BBs and \textit{Structured Workflow} in a fairly subjective manner. Though we highlight- ed some possible ways of representing them, it remains to be explored what an ideal representation would be. Through this work, we are taking the first steps towards systematically structuring this impactful but scattered field of physically assistive robotics. We hope that our framework will speed up the progress in this domain, bringing us one step closer to caregiving robots that can provide long-term assistance.

%% file: src/acknowledgement.tex
\section{Acknowledgment}
This work was funded by the NSF IIS (\#2132846). We thank Daniel Stabile, Abrar Anwar, Boxin Xu and Advika Kumar for their help with experimental setup and figures.